\documentclass{article}
\usepackage{spconf,amsmath,epsfig}
\usepackage{amsmath,amsfonts}
\usepackage{booktabs}
\usepackage{multirow}

\let\OLDthebibliography\thebibliography
\renewcommand\thebibliography[1]{
  \OLDthebibliography{#1}
  \setlength{\parskip}{0pt}
  \setlength{\itemsep}{0pt plus 0.3ex}
}

\begin{document}\sloppy

\def\x{{\mathbf x}}
\def\L{{\cal L}}

\title{FONT: Flow-guided One-shot Talking Head Generation with Natural Head Motions}
%
\name{Jin Liu$^{1,2}$\thanks{$^{\ast}$Corresponding authors. \newline This research is supported in part by  the National Key Research and Development Program of China (No. 2020AAA0140000), and the National Natural Science Foundation of China (No. 61702502).}, Xi Wang$^{1*}$,  Xiaomeng Fu$^{1,2}$, Yesheng Chai$^{1}$,Cai Yu$^{1,2}$,Jiao Dai$^{1*}$,Jizhong Han$^{1}$}

\address{$^{1}$Institute of Information Engineering, Chinese Academy of Sciences, Beijing, China\\
	$^{2}$School of Cyber Security, University of Chinese Academy of Sciences, Beijing, China\\}

\maketitle

\begin{abstract}
One-shot talking head generation has received growing attention in recent years, with various creative and practical applications. An ideal natural and vivid generated talking head video should contain natural head pose changes. However, it is challenging to map head pose sequences from driving audio since there exists a natural gap between audio-visual modalities. In this work, we propose a Flow-guided One-shot model that achieves NaTural head motions(FONT) over generated talking heads. Specifically, we design a probabilistic CVAE-based model to predict head pose sequences from driving audio and source face. Then we develop a keypoint predictor that produces unsupervised keypoints describing the facial structure information from the source face, driving audio and pose sequences. Finally, a flow-guided occlusion-aware generator is employed to produce photo-realistic talking head videos from the estimated keypoints and source face. Extensive experimental results prove that FONT generates talking heads with natural head poses and synchronized mouth shapes, outperforming other compared methods. 
\end{abstract}

\vspace{-2mm}
\section{Introduction}
\label{sec:intro}

Given one source face and driving audio, one-shot talking head generation aims to synthesize a talking head video with reasonable facial animations corresponding to the driving audio ~\cite{zhang2021flow}. This task receives growing attention since it can be used in a wide range of multimedia applications, e.g. video dubbing, digital avatar animation and short video creation.

Some methods~\cite{chen2019hierarchical, prajwal2020lip} have been proposed to edit the mouth area to achieve lip synchronization. However, they neglect the modeling of head motions, thus generating unnatural talking heads that are far from satisfactory from human observation and practical applications. Therefore, researchers turn to focus on generating talking heads with head pose changes. Recent works~\cite{zhou2021pose, yin2022styleheat} choose to introduce an extra auxiliary pose video that guides the head motion changes in the generated talking heads. This formula limits the generalization since it is tedious to find another pose video in one-shot scenario. Hence, some methods try to predict head pose sequence from driving audio.

It is challenging to map driving audio signal into head pose sequence, since there exists natural gap between visual and audio modalities. A great many works~\cite{chen2020talking, zhang2021flow,wang2021audio2head,wang2022one} are proposed to infer head motions from driving audio and source face. However, they neglect the uncertainty in the head pose prediction task and fail to produce natural head poses. In fact, the mapping from driving audio signal to head pose sequence is inherently a one-to-many problem. In real life, people can behave differently in head poses even speaking the same content. Previous methods adopt deterministic models like LSTM or MLP to perform the task, which fundamentally ignore the uncertainty lying between audio signals and head poses. Furthermore, the lack of facial structure modeling in their generation process also leads to blurry artifacts and poor lip-sync quality.

\begin{figure}[t]
	\centering
	\includegraphics[width=\columnwidth]{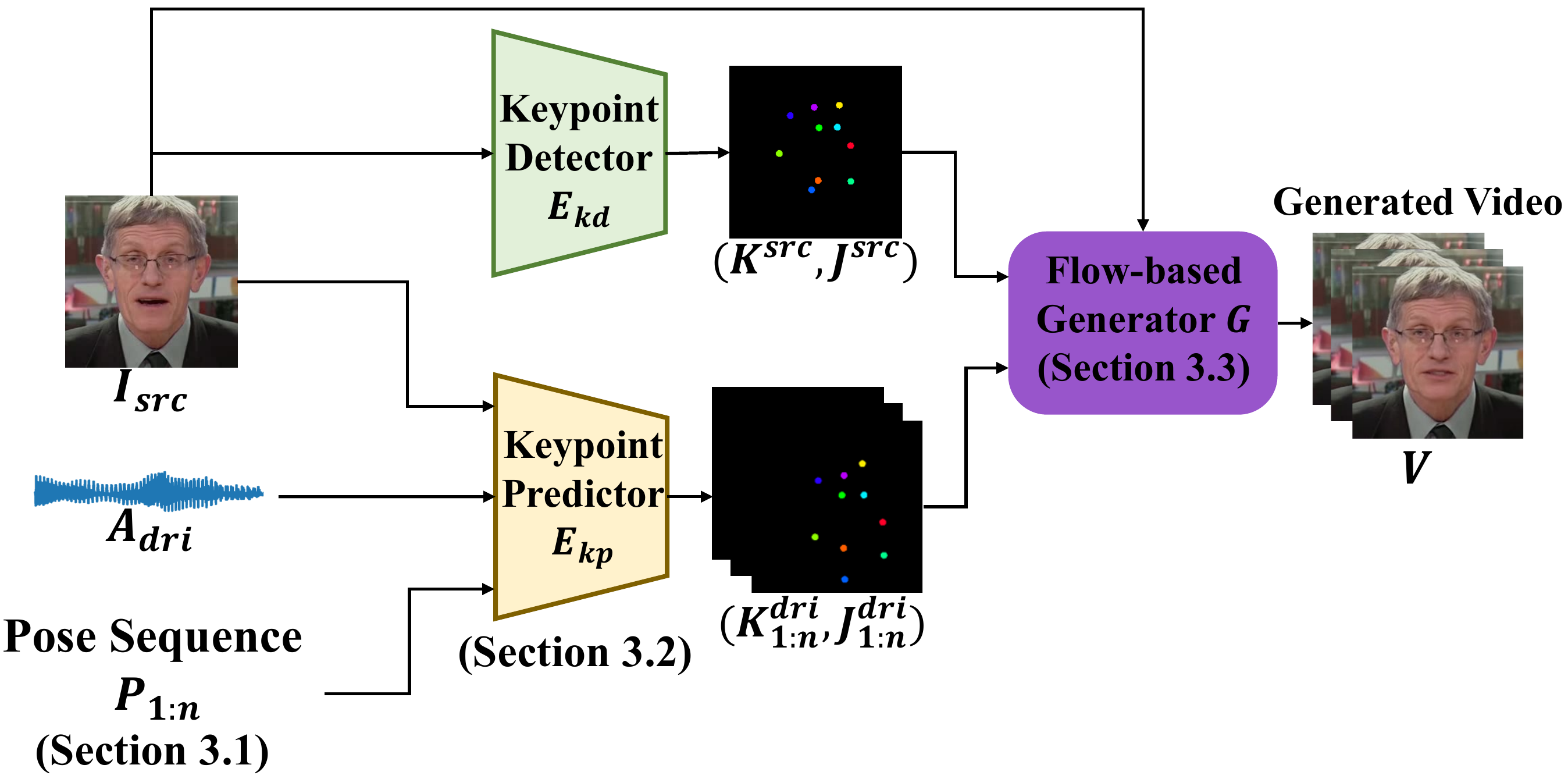} 
	\vspace{-7mm}
	\caption{\textbf{Overview of the proposed method.} }
	\vspace{-3mm}
	\label{fig:overview}
\end{figure}

To solve the above problem, we propose a \textbf{F}low-guided \textbf{O}ne-shot talking head generation network with \textbf{N}a\textbf{T}ural head motions (FONT). The overall framework is shown in Fig.~\ref{fig:overview}. 
The driving pose sequences come from the well-designed head prediction module. Detailedly, a probabilistic CVAE-based network is adopted to generate head pose sequences from driving audio and source face, during which the structural similarity loss is imposed instead of MSE loss. The above operations model the uncertainty and the ambiguous correspondences between audio and head pose modalities, contributing to natural driving head pose sequences. Then inspired by image animation work FOMM~\cite{siarohin2019first}, we predict unsupervised keypoints from the source face, driving audio and poses to model the facial structure location. Finally, the occlusion-aware flow-guided generator produces motion flow to indicate the local facial texture variance and generates new talking heads with natural head poses. Moreover, to improve the lip-sync quality, a pre-trained lip-sync discriminator is utilized during the training process.

Our contributions are as follows: (1) We develop a new flow-guided one-shot talking head generation framework that produces natural head motions. (2) A probabilistic CVAE-based network is designed to generate natural head poses from driving audio and source face. (3) We present a flow-guided occlusion-aware generator to produce keypoint-based motion flow indicating facial structure, thus generating natural talking heads. (4) Extensive experimental results prove that our proposed framework achieves the state-of-the-art level compared to other methods.

\section{Related Work}
\label{sec:related}
\noindent \textbf{One-shot Talking Head Generation.} 
One-shot talking head generation~\cite{chen2020comprises} has long been a significant research topic in the computer vision field. Speech2Vid~\cite{chung2017you} generates talking faces via an encoder-decoder structure and a refinement module. DAVS~\cite{zhou2019talking} and ATVG~\cite{chen2019hierarchical} further improve the quality using disentangled audio-visual representation and external structural information guidance. Wav2Lip~\cite{prajwal2020lip} applies a pre-trained lip-sync discriminator to improve the generation results. Nevertheless, the above methods merely edit the mouth area and leave other facial regions unchanged, producing unnatural and less realistic talking head videos. 

Full-frame talking head generation produces new facial areas but also the neck part of the person, together with the background. MakeitTalk~\cite{zhou2020makelttalk} predicts content and speaker-aware displacement on facial landmarks to guide the talking face generation process. To improve the realness, some methods focus on talking heads with natural head poses~\cite{zhou2021pose, wang2021audio2head}. However, their lack of face structural modeling and the mouth shape constraint causes identity mismatch and poor lip synchronization performance. However, FONT utilizes motion flow as facial structure information and the lip-syn discriminator to solve the above problem.

\noindent \textbf{Head Pose Control.} 
Since there is no explicit head pose information contained in the driving audio signal, it is challenging to achieve head pose control and generate talking heads with natural head motions. Early methods~\cite{prajwal2020lip,zhou2020makelttalk} focus on mouth shape accuracy and produce almost still talking heads. Later, PC-AVS~\cite{zhou2021pose} first propose to rely on auxiliary pose video to obtain head pose guidance. It limits the generalization of this task since obtaining a long pose video is cumbersome in the one-shot scenario. Several methods turn to infer pose sequences directly from audio. Audio2Head ~\cite{wang2021audio2head} and AVCT~\cite{wang2022one} designs a motion-aware LSTM-based network to predict head motions, while HDTF~\cite{zhang2021flow} utilizes Multilayer Perceptron to predict head pose coefficients in morphable face model~\cite{deng2019accurate}. However, the correspondence between audio and head poses contains uncertainty and predicting head poses from audio is actually an ill-posed problem. In real life, people may act different poses even speaking the same content. Hence, instead of utilizing \textit{deterministic} models like other methods, we choose the \textit{probabilistic} CVAE-based~\cite{doersch2016tutorial} network to model the uncertainty in pose generation.

\section{Methodology}
\label{sec:method}

The overview of the proposed method is shown in Fig. \ref{fig:overview}. Driving pose sequence will be predicted first. Then the source face $I_{src}$, driving audio and driving pose sequence are fed into the Keypoint Predictor $E_{kp}$ to predict unsupervised driving keypoints. Then the driving keypoints and source keypoints from $I_{src}$ are taken as inputs to the Generator and produce final talking head videos.

\subsection{Head Pose Prediction Module}

To generate talking heads with natural head motions, the natural head pose sequences should be predicted first. 
Different from previous work~\cite{chen2020talking,wang2021audio2head,wang2022one} which adopt deterministic models like LSTM and traditional GAN to generate pose sequences, we design a VAE-based probabilistic model inspired by CVAE~\cite{doersch2016tutorial}. Pose generation is actually an uncertain ill-posed problem since people may behave differently when speaking the same corpus. Hence, the probabilistic model is more suitable for this task.

\begin{figure}[t]
	\centering
	\includegraphics[width=0.95\columnwidth]{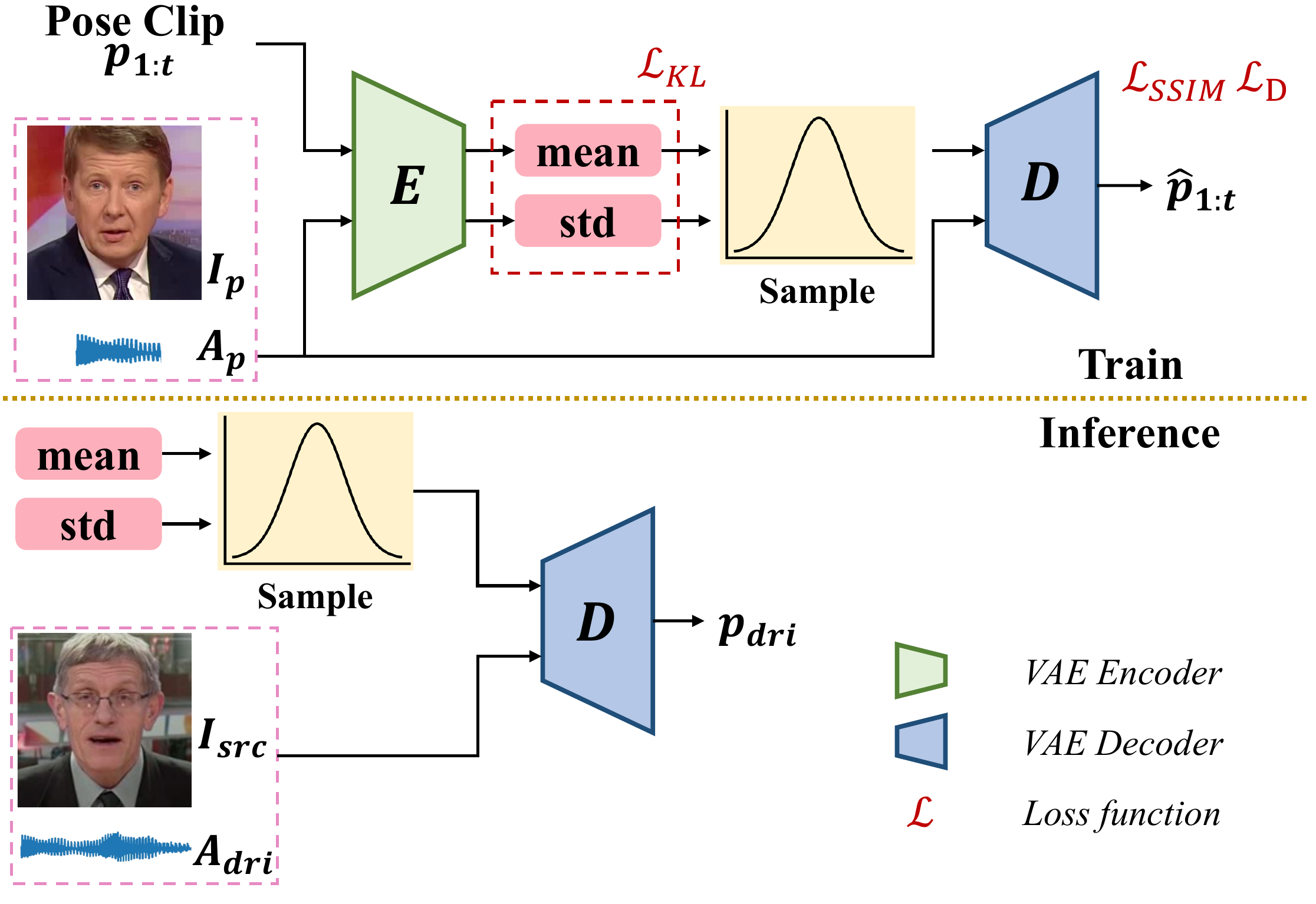} 
		\vspace{-4mm}
	\caption{\textbf{Overview of head pose prediction module.}  }
	\label{fig:head_pose_prediction}
	\vspace{-2mm}
\end{figure}

The head pose prediction module of FONT is shown in Fig. \ref{fig:head_pose_prediction}. Specifically, a 6-dim vector(i.e., 3 for rotation, 1 for scale and 2 for translation) is adopted as pose information representation for each frame. 
Specifically, during the training stage, we utilize paired pose clip $p_{1:t}$,  corresponding audio $A_p$ and head image $I_p$ as inputs. They will be fed into the encoder to predict mean and standard deviation values, which will be later used for re-parametrization. Finally, the sampled data, $I_p$ and $A_p$ are passed into the decoder to predict pose clip $\hat{p}_{1:t}$. The face image and audio are served as the condition information to guide the generation of pose sequence. It is noteworthy that we learn the difference of poses compared to the first frame in $p_{1:t}$ instead of pose itself. This setting eliminates the influence of the various initial head poses in different pose clips. 

As for the loss constraints, commonly used reconstruction loss like MSE loss is not suitable for pose generation, since the task is actually an ill-posed one-to-many mapping problem. Therefore, we utilize the Structural Similarity ~\cite{wang2004image} to keep the consistency between the generated and ground truth pose sequence:
\begin{equation}
	\mathcal{L}_{\text {SSIM }}=1-\frac{\left.\left(2 \mu \hat{\mu}+C_1\right)\left(2 { cov}+C_2\right)\right)}{\left.\left(\mu^2+\hat{\mu}^2+C_1\right)\left(\sigma^2+\hat{\sigma}^2+C_2\right)\right)}.
\end{equation}
$\hat{\mu}$ and $\hat{\sigma}$ are mean and standard deviation of generated pose sequence while ${\mu}$ and ${\sigma}$ are that of the ground truth pose. $cov$ is the covariance between two sequences and $C$ is the constants to stabilize the division. Meanwhile, to guarantee the similarity between latent space distribution and Gaussian distribution, we define $	\mathcal{L}_{KL}$ as the KL-Divergence between the above two distributions. Furthermore, the discriminator is also adopted to improve the realness of the generated pose.
\begin{equation}
	\mathcal{L}_D=\log D\left(p_{gt}\right)+\log \left(1-D\left(p\right)\right).
\end{equation}
The overall loss of pose generation is defined by the combination of $	\mathcal{L}_{\text {SSIM }}$, $	\mathcal{L}_D$ and $	\mathcal{L}_{KL}$.

During inference, driving audio will be divided into several audio clips. They will be fed into the decoder along with $I_{src}$ and sampled latent data to produce pose clips. Finally, the pose clips will be stacked together in chronological order and added to the initial head pose to form the driving pose sequence $P_{1:n}$.

	\vspace{-3mm}
\subsection{Keypoint Predictor}
Inspired by the widely used image animation work FOMM~\cite{siarohin2019first}, we choose to use the unsupervised keypoints and their first order dynamics as the structure representation. 

As illustrated in Fig.~\ref{fig:overview}, the Keypoint Predictor $E_{kp}$ takes source face $I_{src}$, driving audio $A_{dri}$ and predicted pose sequence $p_{1:n}$ as inputs. $E_{kp}$ first utilized three different encoders to extract the corresponding information. Then the three features are combined and fed into the LSTM-based decoder to recurrently predict the corresponding unsupervised structure representation. At each time step $t$, the representation contains learned keypoints $K_{t} \in \mathbb{R}^{N \times 2}$ and the corresponding first order dynamics, i.e. jacobians  $J_{t} \in \mathbb{R}^{N \times 2 \times 2}$ which describes the local affine transformation in the neighborhood area around each keypoint. As for the initial source structure representation, the pre-trained keypoint detector $E_{kd}$ from FOMM~\cite{siarohin2019first} is utilized to provide accurate initial keypoints and first order dynamics. The whole procedure is formulated as:
\begin{equation}
	\begin{aligned}
			(K^{src}, J^{src}) & = E_{kd}(I_{src}),\\
		(K^{dri}_{1:n}, J^{dri}_{1:n}) & = E_{kp}(I_{src}, A_{dri}, p_{1:n}).
	\end{aligned}
\end{equation}
For the training loss, we regard the $ E_{kd}$ as a teacher network and hope $ E_{kp}$ to learn the knowledge of visual structure representation contained in pre-trained $ E_{kd}$. We further define the motion representation of the corresponding ground truth video frame extracted by $ E_{kd}$ as supervision., i.e. $	(K^{gt}, J^{gt})$.    Therefore, the loss term of $ E_{kp}$ is as follows: 
\begin{equation}
	\mathcal{L}_{k p}=\frac{1}{N} \sum_{i=1}^N\left(\left\|K_i^{kp}-K_i^{gt}\right\|_1+\left\|J_i^{kp}-J_i^{gt}\right\|_1\right).
\end{equation}
In this way, the source and driving structure representation are both successfully obtained.

\begin{figure}[t]
	\centering
	\includegraphics[width=\columnwidth]{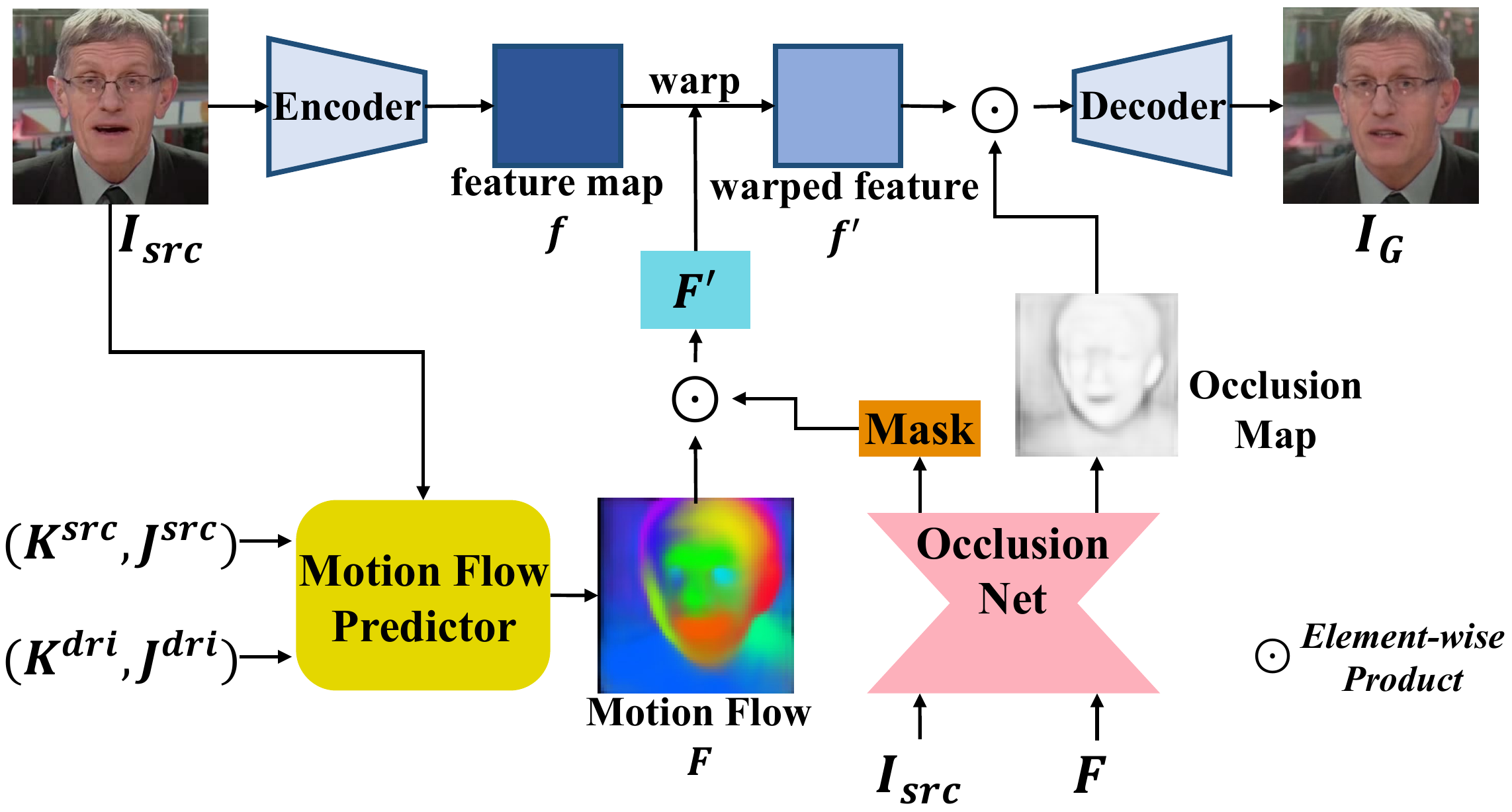} 
		\vspace{-7mm}
	\caption{\textbf{Overview of the flow-guided generator.} }
	
	\label{fig:generator}
	\vspace{-2mm}
\end{figure}

\subsection{Flow-guided Generator}
As shown in Fig.~\ref{fig:generator}, the flow-guided generator produces talking head $I_G$ given $I_{src}$, source and driving structure representation. It mainly contains the motion flow predictor, the occlusion net, the image encoder and decoder. The motion flow predictor first predicts motion flow $F$ indicating the variation in each part of the face from source to driving. Then $I_{src}$ and $F$ are fed into the Occlusion Net to predict the flow mask and occlusion map. The masked motion flow $F^{'}$ is utilized to warp the feature map $f$ of $I_{src}$ to obtain warped feature $f^{'}$. Finally, the occluded feature is sent to the decoder to produce talking head $I_G$. In this way, the decoder obtain the source face texture, motion variance and different confidences among the feature map, which all contribute to the accurate generation process. The encoder and decoder consists of several  convolutional and up-sampling layers. The occlusion net is based on the hourglass net while the motion flow predictor relies on the numerical calculation between two structure representations. During training, the perceptual loss is utilized between the generated frame $I_G$ and ground truth frame ${I}_{gt}$:

\begin{figure}[t]
	\centering
	\includegraphics[width=\columnwidth]{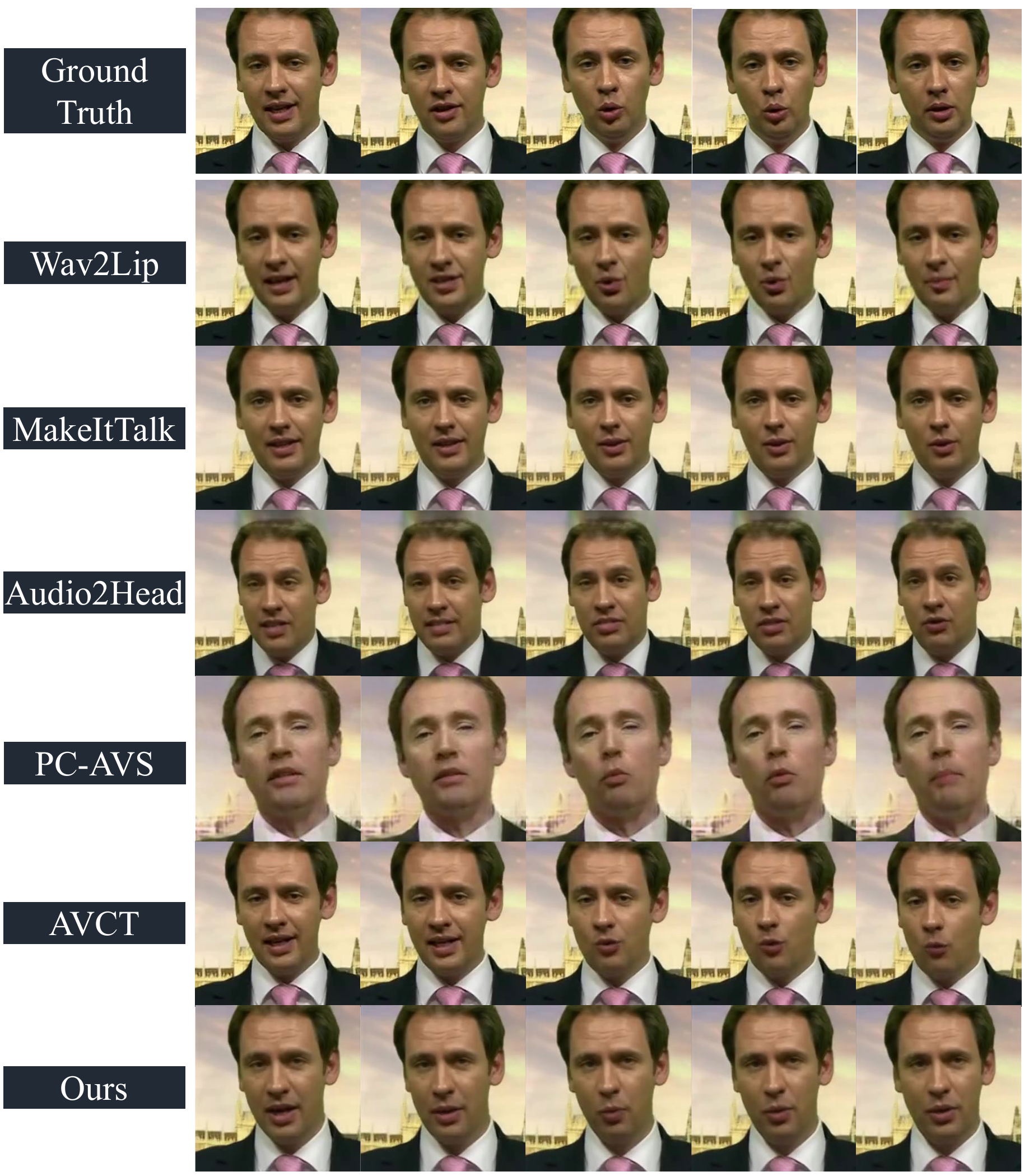} 
		\vspace{-7mm}
	\caption{{Qualitative Comparison with other methods.}    }
	\label{fig:quality}
\end{figure}

\begin{table}[t!]
	\caption{{Quantitative comparisons on LRW dataset. The {bold} and {underlined} notations represents the Top-2 results.  }}
	\centering
	\resizebox{\columnwidth}{!}{
		\begin{tabular}{c c c c c}
			\toprule
			Method & SSIM $\uparrow$ & CPBD $\uparrow$ & LMD $\downarrow$  & LSE-C $\uparrow$   \\
			\midrule
			Wav2Lip      & \underline{0.812}  & 0.172  &5.73     &\textbf{7.237}  \\
			MakeItTalk    & 0.796  & 0.161 &7.13 &3.141  \\
			Audio2Head    &0.743   &0.168  &7.34  &2.135  \\
			PC-AVS         & 0.778  &\underline{0.185}  &3.93    &6.420 \\
			AVCT         &0.805   & 0.181  & \underline{3.56}  &6.567 \\
			Ground Truth     &1.000   &{0.189}  &0.00     &6.876\\
			Ours             &\textbf{0.825}   & \textbf{0.187}  &\textbf{3.48}    & \underline{6.572} \\
			\bottomrule
		\end{tabular}
	}
	\label{tab:LRW}
\end{table}

\vspace{-4mm}
\begin{equation}
	\mathcal{L}_{p e r}=\sum_{i=1}^l || \operatorname{VGG}_i ({I}_{G})-\operatorname{VGG}_i\left(\boldsymbol{I}_{gt}\right) ||_1,
\end{equation}
where $VGG(\cdot)$ denotes the $i_{th}$ channel feature of the pre-trained VGG network. Furthermore, to improve the lip-sync quality, we adopt a pre-trained discriminator to predict the embedding of corresponding audio and video. The discriminator~\cite{prajwal2020lip} is trained to judge the synchronization between randomly sampled audio-visual pairs. We adopt the cosine-similarity between audio and video embedding $a$ and $v$ extracted by the discriminator as the lip-sync loss to indicate the probability of whether the pair is in-sync.

\begin{equation}
\mathcal{L}_{sync}=\frac{v \cdot a}{\max \left(\|v\|_2 \cdot\|a\|_2, \epsilon\right)}
\end{equation}

\begin{figure}[t]
	\centering
	\includegraphics[width=0.8\columnwidth]{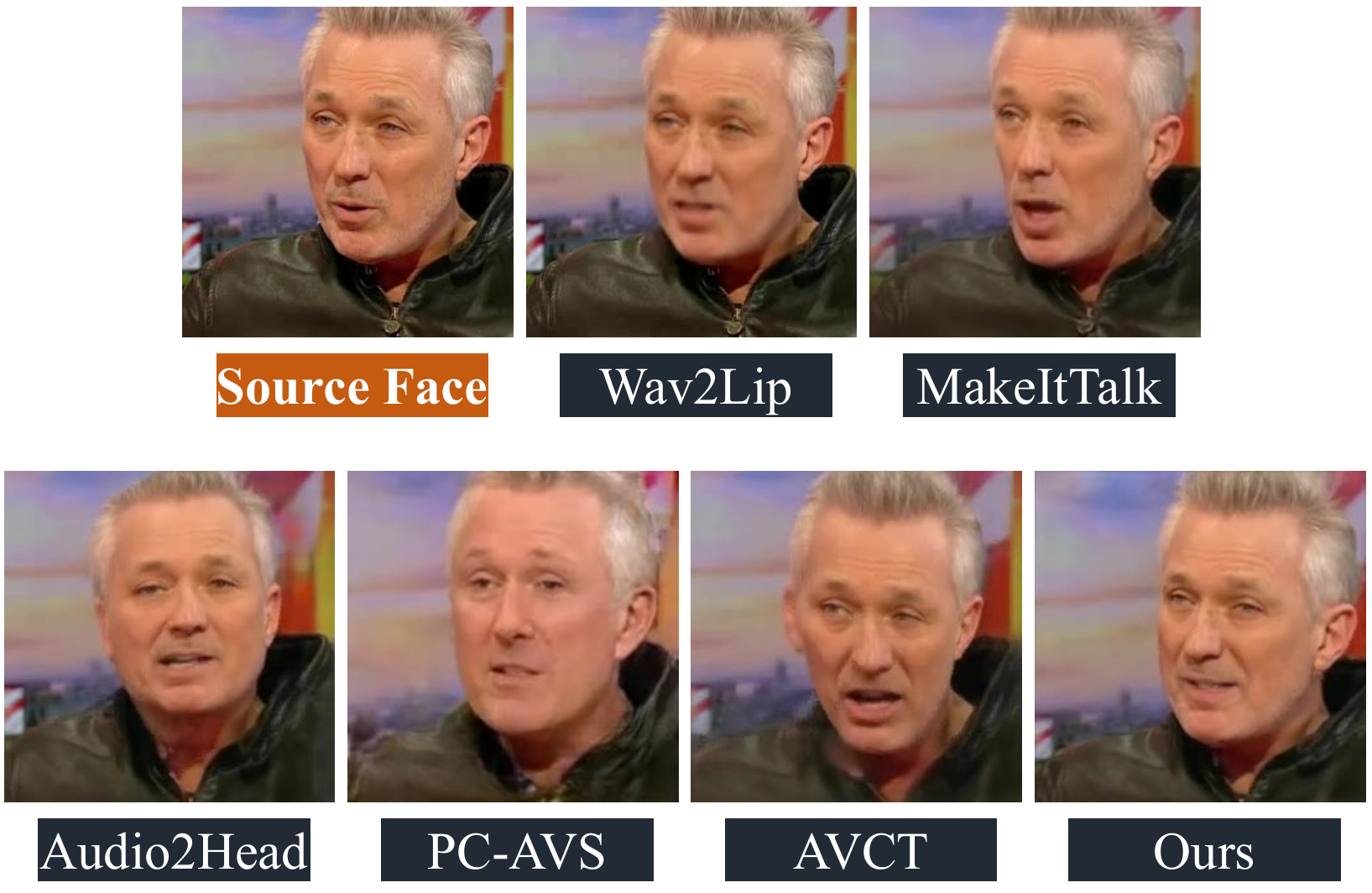} 
		\vspace{-3mm}
	\caption{{Large-pose qualitative comparison results.}    }
	\label{fig:quality_pose}
	\vspace{-2mm}
\end{figure}

\vspace{-2mm}
\begin{table}[t]
	\caption{{Quantitative comparisons on HDTF dataset.  }}
	\centering
	\resizebox{\columnwidth}{!}{
		\begin{tabular}{c c c c c}
			\toprule
			Method & SSIM $\uparrow$ & CPBD $\uparrow$ & LMD $\downarrow$  & LSE-C $\uparrow$   \\
			\midrule
			Wav2Lip      &  \underline{0.786} &  \textbf{0.176} &2.89     &6.97  \\
			MakeItTalk    & 0.751  & 0.132  &5.46     &4.87  \\
			Audio2Head    & 0.735  & 0.145  &4.83    &3.90  \\
			PC-AVS         &  0.762 & 0.164  &3.18     &\underline{7.18}  \\
			AVCT         & 0.769  & 0.167  &\underline{2.71}     &7.09  \\
			Ground Truth     &1.000   &{0.181}  &0.00     &8.58\\
			Ours             &\textbf{0.789}   & \underline{0.169}  &\textbf{2.69}    & \textbf{7.22} \\
			\bottomrule
		\end{tabular}
	}
	\label{tab:HDTF}
	
\end{table}

\section{Experiments}
\label{sec:exp}

\subsection{Experimental Settings}

\noindent \textbf{Datasets.}
We evaluate our method on LRW ~\cite{chung2016lip} and HDTF~\cite{zhang2021flow} datasets. The LRW dataset contains over 1000 short utterances of each 500 different words and all the videos are extracted from BBC television in the wild. The HDTF dataset is a large in the wild audio-visual dataset that consists of long utterances of over 300 subjects.

\noindent \textbf{Implementation Details.}
The face video frames are cropped to $256\times256$ size at 25 FPS and the audio is pre-processed into 16kHz. We compute 28-dim MFCC feature with a window size of 10ms to produce a $28 \times 12$ feature for each video frame.

As for training, our method is trained in stages. The generator is trained with keypoint predictor after the latter gets stable results.  The ADAM optimizer is adopted with an initial leaning rage as $2 \times 10^{-4}$, which linearly decreases to $2 \times 10^{-5}$. We train our model on 1 Tesla V100 GPU and each part requires 0.5, 2 and 3 days for training respectively.

\subsection{Experimental Results}
\noindent \textbf{Evaluation Metrics.}
The performance is evaluated on image quality and lip-sync quality. The SSIM ~\cite{wang2004image} and Cumulative Probability of Blur Detection (CPBD)~\cite{narvekar2009no} scores are utilized to judge the quality of talking head frames. For lip-sync quality, the Landmark Distance(LMD) and Lip-Sync Error-Confidence(LSE-C) are applied. LMD means the average Euclidean distance between corresponding facial landmarks. LSE-C is the confidence score of the correspondence between audio and video features extracted from pre-trained SyncNet~\cite{chung2016out}.

\noindent \textbf{Quantitative Results.}
We choose several state-of-the-art methods as comparison, i.e. Wav2Lip~\cite{prajwal2020lip}, MakeItTalk~\cite{zhou2020makelttalk}, Audio2Head~\cite{wang2021audio2head}, PC-AVS~\cite{zhou2021pose} and AVCT~\cite{wang2022one}. The frames of each method are generated using their official code. The head poses of Wav2Lip and PC-AVS are fixed since they can not obtain head poses from audio. The Ground Truth results are also added for better comparison. 

Detailed results on LRW and HDTF can be found in Table~\ref{tab:LRW} and Table~\ref{tab:HDTF}. FONT achieves the best performance under most of the evaluation metrics on both datasets. As Wav2Lip merely edits the mouth area, it achieves better CPBD score on HDTF. Furthermore, as mentioned in by PC-AVS~\cite{zhou2021pose}, the leading LSE-C only means that Wav2Lip is comparable to the ground truth, not better. The LMD score also proves high-level lip-synchronization of our method. Overall, the above results prove that FONT generates high-quality talking heads.

\begin{figure}[t]
	\centering
	\includegraphics[width=0.95\columnwidth]{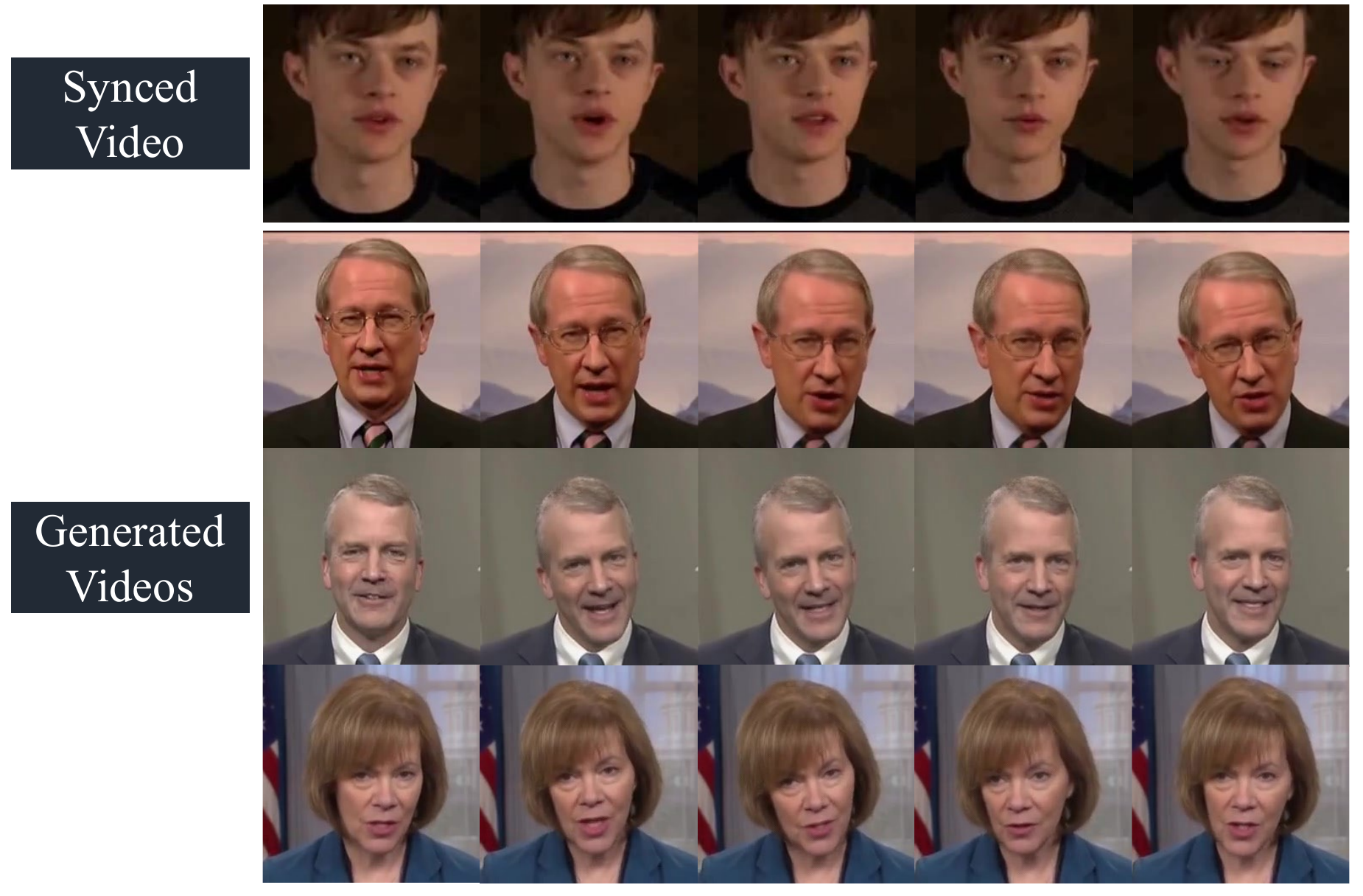} 
		\vspace{-4mm}
	\caption{Qualitative results driven by the same audio and different source faces on HDTF.   }
	\label{fig:quality_hdtf}
\end{figure}

\noindent \textbf{Qualitative Results.}
The qualitative comparison results are shown in Fig.~\ref{fig:quality}. All the frames are generated using the same source face and driving audio. It indicates that FONT generates talking heads with natural head motions, accurate mouth shape and identity information. Specifically, Wav2Lip generates fixed faces and blurry mouth areas. Though MakeitTalk and Audio2Head produce head pose changes, they fail to preserve the lip synchronization corresponding to driving audio. PC-AVS can not preserve the identity information of the source face compared to ground truth. AVCT produces obvious visual artifacts in the background area and sometimes fails to produce an accurate mouth shape. 

We also show comparison results in large-pose faces, as shown in Fig.~\ref{fig:quality_pose}. Other methods displays wired facial shape change and obvious identity mismatch problem, while FONT generates natural head motions while obtaining high-level image quality. Furthermore, Fig.~\ref{fig:quality_hdtf} displays the qualitative results of FONT on the HDTF dataset. It displays the synced video that provides driving audio and generated talking head videos under different source faces.
The results indicate that HDTF produces natural head motions while maintaining high-level lip-syn quality.Please see dynamic demos in the supplementary materials for better comparison.

\noindent \textbf{Ablation Results.}
To evaluate the performance of each component in FONT, we conduct the ablation study on several variants: (1) replace the probabilistic VAE-based model with the deterministic LSTM-based model in head pose generation(\textbf{w/o VAE}),  (2) replace the SSIM loss into the traditional L1 loss (\textbf{w/o $\mathcal{L}_{\text {SSIM }}$}) and (3) remove the lip-sync discriminator from the generator (\textbf{w/o $D_{sync}$}). The results are shown in Table~\ref{tab:ablation}.
Given that the SSIM relates to image pixel accuracy, pose accuracy and image quality become worse when removing the above module. As all the variants share basically the same flow-guided generation pattern, most of them achieve similar CPBD scores. The model \textbf{w/o $D_{sync}$} show a poor LSE-C score indicating bad lip synchronization. The model \textbf{w/o VAE} and \textbf{w/o $\mathcal{L}_{\text {SSIM }}$} fail to produce natural head pose, leading to bad LMD score.

\begin{figure}[t]
	\centering
	\includegraphics[width=0.95\columnwidth]{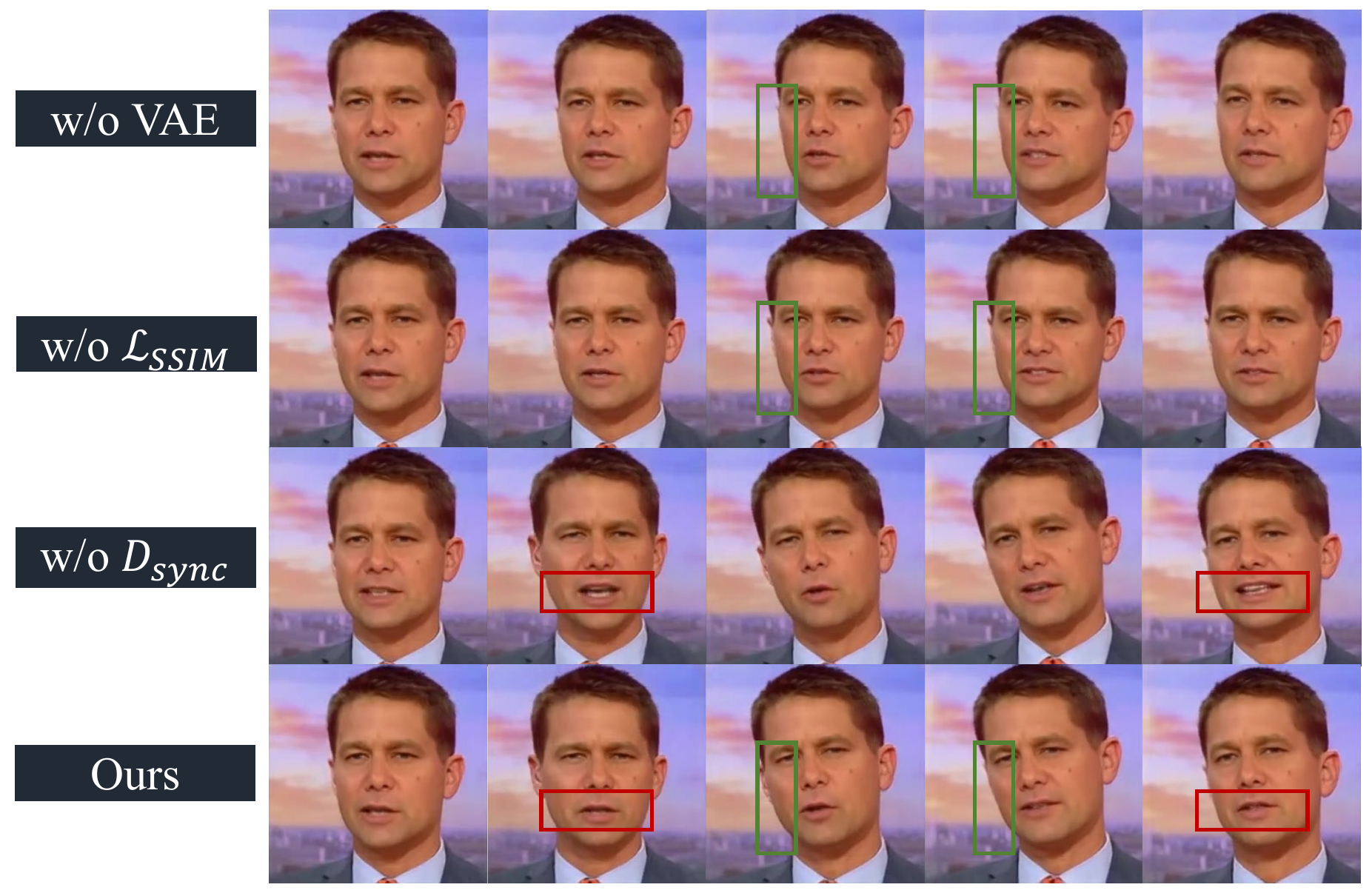} 
		\vspace{-4mm}
	\caption{Qualitative ablation study results. The red and green rectangles indicate the difference of mouth shape and head motion, respectively.   }
	\label{fig:ablation}
\end{figure}

Moreover, we show qualitative ablation results in Fig.~\ref{fig:ablation}. The red and green rectangles mark the difference between each generated frame.
The model \textbf{w/o VAE} and \textbf{w/o $\mathcal{L}_{\text {SSIM }}$} fail to produce dynamic natural head motions and tend to produce average still talking heads. Without $D_{sync}$, the mouth shape accuracy also decreases, as the red rectangle shows. Overall, we see the contribution of each component in FONT.

\begin{table}[t]
	\caption{{Numerical ablation study results.  }}
	\centering
	\resizebox{\columnwidth}{!}{
		\begin{tabular}{c c c c c}
			\toprule
			Method & SSIM $\uparrow$ & CPBD $\uparrow$ & LMD $\downarrow$  & LSE-C $\uparrow$   \\
			\midrule
			w/o VAE   & 0.746  & 0.160  & 5.48   & 7.18 \\
			w/o $\mathcal{L}_{\text {SSIM }}$         & 0.738 & 0.158  & 6.72   & 6.79  \\
			w/o $D_{sync}$         & 0.752  & 0.166  &  3.46  &4.28 \\
			Ours             &\textbf{0.789}   & \textbf{0.169}  &\textbf{2.69}    & \textbf{7.22} \\
			\bottomrule
		\end{tabular}
	}
	\label{tab:ablation}
\end{table}

\section{Conclusion}
In this paper, we present FONT, a flow-guided one-shot model that generates talking heads with natural head motions. The head pose sequence is first predicted by a well-designed probabilistic VAE-based model. After getting the driving pose sequence, we utilize self-supervised keypoints to predict motion flow as face structure representation from the source face and driving audio. Finally, the occlusion-aware flow-guided generator produces talking heads. Both quantitative and qualitative experiments demonstrate that we obtain talking heads with natural poses and high-level lip-sync quality compared with other methods.

For ethical considerations, FONT is intended for the video editing industry and focuses on world-positive use cases and applications. We believe the proper usage of this technique will enhance the development of artificial intelligence research and relevant multimedia applications. To ensure proper use, we will release our codes and contribute to deepfake detection research.

\bibliographystyle{IEEEbib}

\bibliography{ICME2023}
\end{document}